\documentclass[lettersize,journal]{IEEEtran}
\usepackage{amsmath,amsfonts}
\usepackage{algorithmic}
\usepackage{algorithm}
\usepackage{array}
\usepackage[caption=false,font=normalsize,labelfont=sf,textfont=sf]{subfig}
\usepackage{textcomp}
\usepackage{stfloats}
\usepackage{url}
\usepackage{verbatim}
\usepackage{graphicx}
\usepackage{threeparttable}
\usepackage{enumitem}
\usepackage{pgfplots}
\usepackage{amssymb}
\usepackage{multirow}
\usepackage{multicol}
\usepackage{autobreak}
\usetikzlibrary{patterns}
\usepackage[numbers,sort&compress]{natbib}

\begin{document}

\title{P-Transformer: Towards Better Document-to-Document Neural Machine Translation}

\author{Yachao~Li,
        Junhui~Li,
        Jing~Jiang,
        Shimin~Tao,
        Hao~Yang,
        Min~Zhang
\thanks{Yachao~Li and Jing Jiang are with the Key Laboratory of China's Ethnic Languages and Information Technology of Ministry of Education, Northwest Minzu University, Lanzhou 730030, China (e-mail: harry\_lyc@foxmail.com, jiangj0723@163.com)}
\thanks{Junhui~Li and Min~Zhang are with the School of Computer Science and Technology, Soochow University, Suzhou 215006, China (e-mail: \{lijunhui, minzhang\}@suda.edu.cn)}
\thanks{Shimin~Tao and Hao~Yang are with the Huawei Translation Services Center, Beijing, China (e-mail: \{taoshimin,yanghao30\}@huawei.com)}
\thanks{Yachao Li and Jiang Jiang are supported by the National Natural Science Foundation of China (Grant No. 62266038), Junhui Li and Min Zhang are supported by the NSFC (Grant No. 6203600).}
\thanks{This manuscript is under review.}
}
\markboth{Journal of \LaTeX\ Class Files,~Vol.~14, No.~8, August~2021}%
{Shell \MakeLowercase{\textit{et al.}}: A Sample Article Using IEEEtran.cls for IEEE Journals}


\maketitle

\begin{abstract}
Directly training a document-to-document (Doc2Doc) neural machine translation (NMT) via Transformer from scratch, especially on small datasets usually fails to converge. Our dedicated probing tasks show that 1) both the absolute position and relative position information gets gradually weakened or even vanished once it reaches the upper encoder layers, and 2) the vanishing of absolute position information in encoder output causes the training failure of Doc2Doc NMT. To alleviate this problem, we propose a position-aware Transformer (P-Transformer) to enhance both the absolute and relative position information in both self-attention and cross-attention. Specifically, we integrate absolute positional information, i.e., position embeddings, into the query-key pairs both in self-attention and cross-attention through a simple yet effective addition operation. Moreover, we also integrate relative position encoding in self-attention. The proposed P-Transformer utilizes sinusoidal position encoding and does not require any task-specified position embedding, segment embedding, or attention mechanism. Through the above methods, we build a Doc2Doc NMT model with P-Transformer, which ingests the source document and completely generates the target document in a sequence-to-sequence (seq2seq) way. In addition, P-Transformer can be applied to seq2seq-based document-to-sentence (Doc2Sent) and sentence-to-sentence (Sent2Sent) translation. Extensive experimental results of Doc2Doc NMT show that P-Transformer significantly outperforms strong baselines on widely-used 9 document-level datasets in 7 language pairs, covering small-, middle-, and large-scales, and achieves a new state-of-the-art. Experimentation on discourse phenomena shows that our Doc2Doc NMT models improve the translation quality in both BLEU and discourse coherence. We make our code available on Github.\footnote{https://github.com/liyc7711/doc2doc}
\end{abstract}


\section{Introduction}
Document-level neural machine translation, which aims to make the translation more coherent and fluent between sentences, has recently received increasing attention. However, due to the challenge of modeling/generating an input/output document as a single sequence, most related studies still translate a document sentence by sentence and propose various context-aware document-to-sentence models\footnote{In this paper, we follow the notations in \citet{sun-etal-2022-rethinking}, where Doc2Sent models take partial or entire source document as context and generate target-side sentences independently while Doc2Doc models view the source document and target document as long sequences.} (\cite{zhang-2018-doc,junczys:2019,maruf:2019-doc,voita:2019-doc}, to name a few). In these Doc2Sent models, document-level context is usually encoded with an extra-encoder and is properly integrated into a sentence-level translation model. Therefore, such context-aware Doc2Sent models are strictly sentence-level and suffer from limitations caused by not encoding current sentences and their context in the same encoding module~\cite{bao-etal-2021-g,sun-etal-2022-rethinking}. Moreover, independently generating target sentences in a document hinders the usage of target-side document-level context. Different from these studies, in this paper our goal is to build a document-to-document translation model which encodes a document as a single unit and generates its translation as well as a single unit.

Although extending the translation unit from a single sentence to multiple sentences (e.g., 4) achieves good performance~\cite{tiedemann:2017,agrawal:2018,ma:2020-simple}, directly training a seq2seq-based Doc2Doc model, i.e., Transformer~\cite{vaswani_etal:17} from scratch tends to fail~\cite{liu-2020-mbart,bao-etal-2021-g}, especially on small training data.\footnote{We note that a Doc2Doc Transformer can be successfully trained upon a pre-trained model (e.g., mBART) or on large-scale training data. However, this requires additional large-scale datasets.} For example,~\citet{bao-etal-2021-g} show that the failure is due to the local minima during training. As a result, the attention weights in the cross-attention between the encoder and the decoder are flat, with large entropy values. Based on the observation of attention weight distribution, \citet{bao-etal-2021-g} propose G-Transformer, the first Doc2Doc NMT model by introducing local bias to attention. Later, \citet{sun-etal-2022-rethinking} observe that vanilla Transformer~\cite{vaswani_etal:17} with appropriate data augmentation techniques can achieve good performance for Doc2Doc translation, which alleviates the issues of limited training data. Although both the studies \cite{bao-etal-2021-g,sun-etal-2022-rethinking} successfully adapt Transformer to properly cater to long-sequence input and output, they have not answered the behind reasons for the failure when directly training a seq2seq-based Doc2Doc model on vanilla Transformer. Moreover, strictly forcing sentence-to-sentence alignment between the input and the output prohibits G-Transformer~\cite{bao-etal-2021-g} from being applied for other seq2seq tasks, like text summarization while MR Doc2Doc~\cite{sun-etal-2022-rethinking} may not able to recover the sentence-to-sentence alignment between a source document and its translation.\footnote{In our re-implementation, 6 out of 115 documents of the English-to-German TED dataset have the different number of sentences in the source and the translation.} 

In this paper, we take a step further to explore the reason that causes the training failure of Doc2Doc NMT with vanilla Transformer model~\cite{vaswani_etal:17}. Similar to~\citet{bao-etal-2021-g}, our Doc2Doc Transformer fails to converge when training on small training data, e.g., English-to-German TED dataset with 10.9K documents and 210K sentences. We conjecture that even with residual connections, the Transformer model could not preserve sufficient position information at the top layer once the input sequence becomes long. As a result, the cross-attention between the encoder and the decoder has little knowledge about the position information of the source-side hidden states and thus results in flat attention weights with large entropy values. To verify the conjecture, we take three dedicated probing tasks to access the position information encoded in the encoder layers (Section~\ref{abl:pos}). As shown in Figure~\ref{fig:probing}, the experimental results reveal that the accuracy of absolute position prediction for the top layer (i.e., Layer 6) of the Doc2Doc model is extremely as low as 0.5\% while the accuracy for the Sent2Sent model is 62.2\%.
Moreover, as shown in Figure~\ref{fig:prob-rel} and Figure~\ref{fig:prob-order}, there is a visible tendency that with the increase of input sequence, the relative position information degrades significantly for both Sent2Sent and Doc2Doc models. That is to say, the position information has almost vanished when encoding a document as a single unit.

Motivated by the results of position information probing tasks, it is essential to enhance the position information of the Transformer model when building Doc2Doc NMT. To this end, we propose an incredibly simple yet effective approach and construct P-Transformer, a Transformer-based model with position-aware attention, in which the query-key pairs are explicitly equipped with their corresponding (absolute) position information. We then apply position-aware attention to both self-attention and cross-attention modules in Transformer. Moreover, we also integrate relative position information into self-attention both on the source and target side. The position-aware attention, in turn, will enhance the position information embedded in the hidden states of the top encoder layer, which is helpful for Doc2Doc translation. Experimental results on 9 popular document-level translation datasets in 7 language pairs show that our proposed P-Transformer significantly outperforms the strong baselines and achieves a new state-of-the-art performance.

Overall, the contributions of this paper are three-fold.
\begin{itemize}
\item For both Sent2Sent and Doc2Doc translation, we provide a probing study to investigate whether the position information gets weakened or even vanished at the top Transformer encoder layer. Through the probing tasks, we find the main reason causing the training failure of Doc2Doc translation with vanilla Transformer.
\item We propose P-Transformer with position-aware attention which could be successfully trained for Doc2Doc translation. Moreover, P-Transformer could also be used to boost the performance of Doc2Sent and Sent2Sent translation.
\item Our extensive experimental results show that the proposed P-Transformer significantly outperforms the strong baselines on 9 document-level datasets in 7 language pairs covering small-, middle- and large-scales. In addition, the P-Transformer achieves state-of-the-art performance.
\end{itemize}

\section{Position Information Probing Tasks}
\label{abl:pos}

Positional encoding plays a crucial role in Transformer to make use of the token order of a sequence. Specifically, the word embeddings (WEs) and position embeddings (PEs) are summed, i.e., $\mathbf{WE} + \mathbf{PE}$ as the word representation, which is fed to Transformer. 
Specifically, vanilla Transformer~\cite{vaswani_etal:17} uses sinusoidal functions to parameterize PEs in a fixed ad hoc way. 
Theoretically, the position information from the input can efficiently be propagated to the upper layers through residual connections. To the best of our knowledge, however, few relevant studies have looked into whether the hidden states, especially in upper layers, can preserve appropriate position information. Therefore, we propose three probing tasks from different views to properly measure how much position information is encoded in Transformer encoder layers. We implement these probing tasks upon the linguistic features probing toolkit of~\citet{conneau_etal:acl18} with default parameter settings.\footnote{\url{https://github.com/facebookresearch/SentEval}} The training set, validation set, and test set are from the popular English-to-German TED document-level dataset (See Section~\ref{sect:data} for data description).\footnote{When we apply the three probing tasks on middle-scale English-to-German Europarl dataset, experimental results show that Doc2Doc Transformer can preserve position information.}

\subsection{Absolute Position Probing Task}

We formalize the absolute position probing task as follows: given the hidden state ${\mathbf{h}_i}\in\mathbb{R}^{D}$ of the $i$-th source-side word, our goal is to predict its absolute position value, i.e., $i$ through a neural network. Inspired by the linguistic feature probing tasks proposed in \citet{conneau_etal:acl18}, we treat it as $K$-label classification problem and train a classifier through a fully-connected network to map a hidden state ${\mathbf{h}}$ into a $K$-class probability distribution via:

\begin{equation}
    P_{A}\left(\mathbf{k}|\mathbf{h}\right) =  \text{Softmax}\left(\mathbf{W_{A}}\mathbf{h}\right),
\end{equation}
where $\mathbf{W_A}\in\mathbb{R}^{K\times D}$ is a trainable matrix, $\mathbf{k}\in\{1, 2, \cdots, K\}$ is an absolute position, $K$ is the maximum input length, and $D$ is the hidden state size. 

\begin{figure}[ht]
	\setlength{\abovecaptionskip}{0pt}
	\begin{center}
		\pgfplotsset{compat=1.13}
		\begin{tikzpicture}
		\tikzset{every node}=[font=\scriptsize]
		\begin{axis}
		[height=4.3cm,width=7.0cm,enlargelimits=0.07, tick align=inside, legend style={cells={anchor=west},legend pos=south west}, xticklabels={Layer~1,Layer~2,Layer~3,Layer~4,Layer~5,Layer~6},
		xtick={1,2,3,4,5,6},
		ylabel={Accuracy (\%)},
		xlabel={Encoder Layer},
		ymax=100,
		ymin=0,
		ytick distance=20,
		]	

		\addplot+[smooth,color=blue,mark color=blue,mark=square*] coordinates
		{(1,99.85) (2,99.51) (3,98.01) (4,91.70) (5,78.48) (6,62.17) };
		\addlegendentry{\scriptsize{Trans. (Sent2Sent)}}

		\addplot+[smooth,color=blue,mark color=blue,mark=star] coordinates
		{(1,99.59) (2,98.44) (3,94.50) (4,76.51) (5,41.03) (6,0.49)};
		\addlegendentry{\scriptsize{Trans. (Doc2Doc)}}

		\end{axis}
		\end{tikzpicture}
		\caption{Accuracy of absolute position probing task for Transformer-based Sent2Sent and Doc2Doc NMT. We take probing on different encoder layers to access the position information encoded. The higher the probing accuracy, the more position information is encoded in the hidden state of this layer.}\label{fig:probing}
	\end{center}
\end{figure}
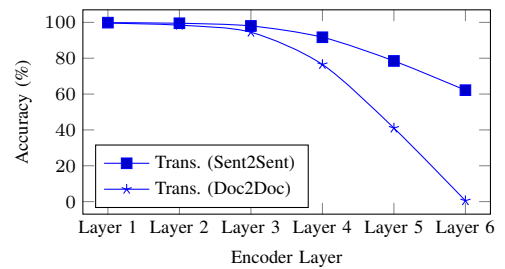

Figure~\ref{fig:probing} compares the classification accuracy for hidden states in different encoder layers of both Sent2Sent and Doc2Doc NMT. From the figure we have the following observations:

\begin{itemize}
    \item For both Sent2Sent and Doc2Doc NMT, the accuracy is very high for the hidden states of low layers (layer~1 $\sim$ 3), suggesting that position information is strongly preserved. From layer 4 to layer 6, there is an obvious trend that the accuracy starts to decrease, suggesting that the position information gets weakened or even vanished once it reaches the upper layers. 
    \item Let us focus on the accuracy of the top layer, i.e., layer~6 where the hidden states will be used as the final output of the encoder. 
    The accuracy for Sent2Sent NMT is 62.2\% while the accuracy for Doc2Doc NMT is as extremely low as 0.5\%. It indicates that the final output of the encoder for Doc2Doc NMT almost does not contain absolute position information. Therefore, this will confuse the cross-attention in the decoder and result in flat attention weights.
\end{itemize}

Even when we loose the metric to allow approximate matching, i.e., a prediction is correct if the predicted position is in the $\pm 3$ window-size of the correct position, the accuracy of the top layer for Sent2Sent NMT increases to 95.1\% while it is still as low as 1.4\% for Doc2Doc NMT. This further suggests that position information has almost vanished in the final encoder output of Doc2Doc NMT.

\subsection{Relative Position Probing Task}

Given a pair of hidden states ${\mathbf{h}_i},{\mathbf{h}_j}, \in\mathbb{R}^{D}$ ($i\ne j$) of the $i$-th and the $j$-th source-side words, the relative position probing task is to predict their relative position, i.e., $i - j$, between the word pair. Similar to the absolute position probing, we use a neural network to map the two hidden states into a $2K$-class probability distribution via:
\begin{equation}
    P_{R}\left(\mathbf{k}|\mathbf{h_i}, \mathbf{h_j}\right) = \text{Softmax}\left(\mathbf{W_{R}}\left(\mathbf{h_i} - \mathbf{h_j}\right)\right),
\label{eq:prob:rel}
\end{equation}
where $\mathbf{W_{R}}\in\mathbb{R}^{2K\times D}$ is a trainable matrix, $K$ is the maximum relative distance, and $D$ is the hidden state size. $\mathbf{k}\in\{-K, -K\text{+}1, \cdots, -1, 1, \cdots, K\text{-}1, K\}$ is relative pairwise distance. 

Given a sequence $\left(s_1, s_2, \cdots, s_I\right)$ with $I$ words, the total number of word pairs from it is $I\left(I\text{-}1\right)$. Instead of using all word pairs, we sample $I$ word pairs by setting $h_i$ as $h_1$, $h_2$, $\cdots$, and $h_I$, respectively while randomly sampling $h_j$ from $\pm K$ window-size of $h_i$. 
Figure~\ref{fig:prob-rel} shows the accuracy curves of the relative position probing on the hidden states of the top encoder layer, where the maximum relative distance $K=20$.
From Figure~\ref{fig:prob-rel}, we observe that:

\begin{figure}[ht]
	\setlength{\abovecaptionskip}{0pt}
	\begin{center}
		\pgfplotsset{compat=1.13}
		\begin{tikzpicture}
		\tikzset{every node}=[font=\scriptsize]
		\begin{axis}[height=4.3cm,width=7.0cm,enlargelimits=0.03, tick align=inside, legend style={cells={anchor=west},legend pos=north west}, 
		xticklabels={-20,,,,,-15,,,,,-10,,,,,-5,,,,-1,,,,,5,,,,,10,,,,,15,,,,,20},
		xtick={1,2,3,4,5,6,7,8,9,10,11,12,13,14,15,16,17,18,19,20,21,22,23,24,25,26,27,28,29,30,31,32,33,34,35,36,37,38,39,40},
		ylabel={Accuracy (\%)},
		xlabel={Relative Distance},
		ymax=50,
		ymin=0,
		ytick distance=10,
		]
		\addplot+ [smooth,color=blue,mark color=blue,mark=square*] coordinates
		{(1,17.44) (2,11.12) (3,13.77) (4,5.85) (5,6.74) (6,13.02) (7,9.30) (8,9.44) (9,10.41) (10,2.99) (11,12.07) (12,9.40) (13,15.19) (14,11.36) (15,6.99) (16,12.07) (17,16.14) (18,16.32) (19,23.30) (20,44.07) (21,39.75) (22,28.14) (23,14.77) (24,16.00) (25,14.19) (26,11.31) (27,8.78) (28,9.48) (29,13.09) (30,8.35) (31,4.38) (32,10.47) (33,13.47) (34,2.28) (35,4.07) (36,7.91) (37,11.70) (38,14.25) (39,6.55) (40,23.08)};  
		\addlegendentry{\tiny{Trans. (Sent2Sent)}}
		
		\addplot+ [smooth,color=blue,mark color=blue,mark=star] coordinates
		{(1,3.76) (2,1.87) (3,1.02) (4,4.44) (5,1.80) (6,0.07) (7,1.22) (8,1.10) (9,0.06) (10,0.14) (11,1.37) (12,3.39) (13,5.73) (14,0.00) (15,1.27) (16,0.28) (17,1.97) (18,2.16) (19,6.38) (20,36.44) (21,16.20) (22,6.21) (23,0.66) (24,0.75) (25,2.38) (26,0.21) (27,1.14) (28,0.97) (29,0.41) (30,1.71) (31,0.81) (32,1.84) (33,1.19) (34,0.95) (35,0.14) (36,0.74) (37,0.36) (38,1.88) (39,4.05) (40,7.01)};  
		\addlegendentry{\tiny{Trans. (Doc2Doc)}}

		\end{axis}
		\end{tikzpicture}
		\caption{Accuracy of relative position probing task for  Transformer-based Sent2Sent and Doc2Doc NMT. }\label{fig:prob-rel}
	\end{center}
\end{figure}
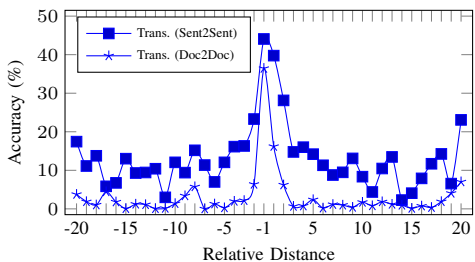

\begin{itemize}
    \item For both Sent2Sent and Doc2Doc NMT, the accuracy is high among hidden states of the local area with a relative distance of $\pm$ 1, suggesting that relative position information is well preserved.
    \item For relative distance from $\pm$ 2 to longer, there is a visible trend that the accuracy starts to decrease, which is more serious in the long sequence input of Doc2Doc NMT. This suggests that the relative position information gets weakened once the word pairs in a long distance.
    \item The probing results indicate that the relative position information is also weak or even vanishing in long sequence Doc2Doc NMT encoder output.
\end{itemize}

\subsection{Word Order Probing Task}

Given a pair of hidden states ${\mathbf{h}_i},{\mathbf{h}_j}, \in\mathbb{R}^{D}$ ($i\ne j$) of the $i$-th and the $j$-th source-side words, the word order probing task is a two-class classification problem by predicting the word order between the word pair, i.e., 0 for $i>j$ and 1 for $i<j$. Similar to the relative position probing task, we use a neural network to map the two hidden states into a two-class probability distribution via:
\begin{equation}
    P_{O}\left(\mathbf{k}|\mathbf{h_i}, \mathbf{h_j}\right) = \text{Softmax}\left(\mathbf{W_{O}}\left(\mathbf{h_i} - \mathbf{h_j}\right)\right),
\label {eq:prob:order}
\end{equation}
where $\mathbf{W_{O}}\in\mathbb{R}^{2\times D}$ is a training matrix, $\mathbf{k}\in\{0, 1\}$, and $D$ is the hidden state size. 

Same as the relative position probing task, we sample $I$ word pairs from $I$-length input sequence by constraining $h_j$ from $\pm K$ window-size of $h_i$. Figure~\ref{fig:prob-order} shows the accuracy curves of the word order probing task on the hidden states of the top encoder layer, where the maximum relative distance $K=20$. From it, we observe that:

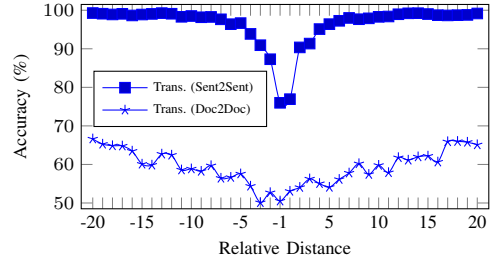
\begin{figure}[ht]
	\setlength{\abovecaptionskip}{0pt}
	\begin{center}
		\pgfplotsset{compat=1.13}
		\begin{tikzpicture}
		\tikzset{every node}=[font=\scriptsize]
		\begin{axis}[height=4.3cm,width=7.0cm,enlargelimits=0.03, tick align=inside, legend style={cells={anchor=west},legend pos=north west,yshift=-0.8cm}, 
		xticklabels={-20,,,,,-15,,,,,-10,,,,,-5,,,,-1,,,,,5,,,,,10,,,,,15,,,,,20},
		xtick={1,2,3,4,5,6,7,8,9,10,11,12,13,14,15,16,17,18,19,20,21,22,23,24,25,26,27,28,29,30,31,32,33,34,35,36,37,38,39,40},
		ylabel={Accuracy (\%)},
		xlabel={Relative Distance},
		ymax=100,
		ymin=50,
		ytick distance=10,
		]
		\addplot+ [smooth,color=blue,mark color=blue,mark=square*] coordinates
		{(1,99.27) (2,99.10) (3,98.91) (4,99.07) (5,98.65) (6,98.89) (7,99.05) (8,99.25) (9,99.09) (10,98.30) (11,98.50) (12,98.18) (13,98.26) (14,97.65) (15,96.38) (16,96.66) (17,93.86) (18,90.93) (19,87.29) (20,75.99) (21,76.93) (22,90.36) (23,91.35) (24,95.11) (25,96.39) (26,97.26) (27,98.00) (28,97.67) (29,97.90) (30,98.29) (31,98.38) (32,98.96) (33,99.20) (34,99.23) (35,99.01) (36,98.72) (37,98.66) (38,98.72) (39,98.79) (40,99.16)};  
		\addlegendentry{\tiny{Trans. (Sent2Sent)}}
		
		\addplot+ [smooth,color=blue,mark color=blue,mark=star] coordinates
		{(1,66.62) (2,65.267) (3,64.86) (4,64.79) (5,63.46) (6,60.08) (7,59.83) (8,62.69) (9,62.42) (10,58.58) (11,58.85) (12,58.23) (13,59.68) (14,56.43) (15,56.65) (16,57.55) (17,54.37) (18,50.10) (19,52.69) (20,50.51) (21,53.05) (22,54.05) (23,56.25) (24,55.05) (25,54.08) (26,56.12) (27,57.78) (28,60.20) (29,57.42) (30,59.78) (31,57.88) (32,61.77) (33,61.12) (34,61.97) (35,62.17) (36,60.63) (37,65.90) (38,65.97) (39,65.77) (40,65.11)};  
		\addlegendentry{\tiny{Trans. (Doc2Doc)}}

		\end{axis}
		\end{tikzpicture}
		\caption{Accuracy of word order probing task for  Transformer-based Sent2Sent and Doc2Doc NMT. }\label{fig:prob-order}
	\end{center}
\end{figure}

\begin{itemize}
    \item For Sent2Sent NMT, the probing accuracy is about 76\% on the hidden states of local distance (-1 or 1), suggesting that word order information is well preserved in the local area. In the relative distance from 2 to the longer sequence, there is a visible trend that the accuracy starts to increase, reaching about 96\% $\sim$ 100\%. This suggests that the word order information is strongly preserved at a long distance. 
    \item For Doc2Doc NMT, the probing accuracy is low (about 50\% $\sim$ 55\%) on the hidden states of local distance~(-5 $\sim$ 5), suggesting that word order information is very weak in the local area. The relative distance over $\pm$ 5 to the longer sequence, there is a trend that the accuracy starts to increase.
    \item The word order information of Sent2Sent NMT encoder output is strongly preserved, while it is very weak in Doc2Doc NMT encoder output. 
\end{itemize}

\section{P-Transformer: Position-Aware Transformer}

In Transformer~\cite{vaswani_etal:17}, the attention function is viewed as a mapping between a query and a set of key-value pairs, to an output. The Transformer applies the attention function in two self-attention modules and one cross-attention module. To enhance position information in the attention function, we propose position-aware attention in which the query-key pairs are explicitly equipped with their corresponding position information. Moreover, we also propose to integrate relative position into the self-attention module to further improve the ability to perceive position information. In the next, we present the position-aware self-attention, cross-attention, and the relative position encoding for self-attention in detail.

\subsection{Position-Aware Self-Attention}
For the self-attention module, we define its input as ${\mathbf{H}} \in {\mathbb{R}^{I \times D}}$, where $I$ is the input length and $D$ is the size of hidden states. The original self-attention~\cite{vaswani_etal:17} computes the input as:

\begin{equation}
\label{equ:attn}
 \text{Softmax}\left(\frac{\left(\mathbf{HW_Q}\right)\left(\mathbf{HW_K}\right)^\mathrm{T}}{\sqrt{D}}\right) \left(\mathbf{HW_V}\right),
\end{equation}
 where ${\mathbf{W_Q}}, {\mathbf{W_K}}, {\mathbf{W_V}}\in\mathbb{R}^{D\times D}$ denote the query, key, and value projection matrix, respectively. 
In Eq.~\ref{equ:attn}, position information is implicitly encoded in the hidden states $\mathbf{H}$.
To enable the self-attention to be aware of the absolute position of the input, we simply add the position embeddings to the hidden states of query-key pairs, which we refer to as \textit{position-aware self-attention}. The updated Eq.~\ref{equ:attn} is computed as:

\begin{equation}
\resizebox{0.86\hsize}{!}{$
\label{equ:attn-self}
\text{Softmax}\left(\frac{(\mathbf{(H+P)W_Q})\left(\mathbf{(H+P)W_K}\right)^\mathrm{T}}{\sqrt{D}}\right) \left(\mathbf{HW_V}\right),
$}
\end{equation}
where ${\mathbf{P}}\in\mathbb{R}^{I\times D}$ is the absolute position embeddings, for which we use the sinusoidal position encoding proposed in~\citet{vaswani_etal:17}. Therefore, the proposed position-aware self-attention does not introduce new parameters.

Adding position information to self-attention is not novel. For example, related studies in \cite{shaw-etal-2018-self,wang-etal-2019-self,yang_nips_2019,chen-etal-2021-simple} propose relative position or more sophisticated position encoding to the self-attention module. Compared to them, our approach is simple enough and does not need any new position encoding mechanism. Moreover, it does not change other parts of the Transformer.

\subsection{Position-Aware Cross-Attention}

In Transformer, cross-attention, also called encoder-decoder attention, passes information from the encoder to the decoder. The original cross-attention \cite{vaswani_etal:17} computes the input as:
\begin{equation}
\label{equ:cross-attn}
\text{Softmax}\left(\frac{\left(\mathbf{H_{s}W_Q}\right)\left(\mathbf{H_{t}W_K}\right)^\mathrm{T}}{\sqrt{D}}\right) \left(\mathbf{H_s}\mathbf{W_V}\right),
\end{equation}
where ${\mathbf{H_s}}\in\mathbb{R}^{I_s \times D}$, ${\mathbf{H_t}}\in\mathbb{R}^{I_t \times D}$ denote the output of the encoder and the output of self-attention of the decoder side, respectively, while $I_s$ and $I_t$ denote the lengths of source-side and target-side sequences, respectively. Similar to position-aware self-attention, we enhance the cross-attention to be aware of the absolute position of the hidden states of the encoder and the decoder, which we refer to as \textit{position-aware cross-attention}. The updated Eq.~\ref{equ:cross-attn} is computed as:

\begin{equation}
\resizebox{0.86\hsize}{!}{$
\label{equ:attn-cross}
\text{Softmax}\left(\frac{(\mathbf{(H_t+P_t)W_Q})\left(\mathbf{(H_s+P_s)W_K}\right)^\mathrm{T}}{\sqrt{D}}\right) \left(\mathbf{H_sW_V}\right),
$}
\end{equation}
where ${\mathbf{P_s}} \in {\mathbb{R}^{I_s \times D}}$, ${\mathbf{P_t}} \in {\mathbb{R}^{I_t \times D}}$ denote the corresponding sinusoidal position embeddings of ${\mathbf{H_s}}$ and ${\mathbf{H_t}}$, respectively.

After adding absolute position information to the query-key pairs, the cross-attention will learn to effectively extract useful context information over the source document by comparing the absolute position of the query against the position of the keys.  
Compared with the methods of integrating local attention and global attention~\cite{longformer:2020,bao-etal-2021-g}, our position-aware cross-attention is much simpler without introducing new parameters.

\subsection{Relative Position for Self-attention}

From the relative position probing result in Figure~\ref{fig:prob-rel}, we find that the relative position information will decrease rapidly for pairs of long distance. To enhance relative position information in self-attention, we follow \citet{huang-2019-music:trans} to integrate the relative position embeddings into the query-key component in the self-attention network.
 
Given an input $\mathbf{H}\in\mathbb{R}^{I\times D}$, the relative position-aware self-attention is computed as:
\begin{equation}
\label{equ:attn-rel}
\text{Softmax}\left(\frac{(\mathbf{HW_Q})\left(\mathbf{HW_K}\right)^\mathrm{T}+\mathbf{S}_{rel}}{\sqrt{D}}\right) \left(\mathbf{HW_V}\right),
\end{equation}

\begin{equation}
\label{equ:rel-emb}
\mathbf{S}_{rel}=(\mathbf{HW_Q})\mathbf{R}^T,
\end{equation}
where $\mathbf{R} \in {\mathbb{R}^{I \times I \times D}}$ is a tensor of relative positional embeddings.\footnote{In implementation, the result of $\mathbf{HW_Q}$ will be reshaped as $\mathbb{R}^{I\times 1\times D}$ before multiplying with $\mathbf{R}^T$.} Note that unlike \citet{shaw-etal-2018-self}, here we do not clip the minimum/maximum relative position (i.e., $\pm 512$, 512 is the maximum sequence length) to a certain value. When we apply relative position for position-aware self-attention, Eq.~\ref{equ:attn-rel} and Eq.~\ref{equ:rel-emb} are updated as:

\begin{equation}
\resizebox{0.86\hsize}{!}{$
\label{equ:attn-rel2}
\text{Softmax}\left(\frac{\left((\mathbf{H}+\mathbf{P})\mathbf{W_Q}\right)\left(\mathbf{(H+P)W_K}\right)^\mathrm{T}+\mathbf{S}_{rel}}{\sqrt{D}}\right) \left(\mathbf{HW_V}\right),
$}
\end{equation}

\begin{equation}
\label{equ:rel-emb2}
\mathbf{S}_{rel}=\left(\mathbf{\left(H+P\right)W_Q}\right)\mathbf{R}^T.
\end{equation}
The relative position~\cite{shaw-etal-2018-self} provides the relative pairwise distance of the input sequence, which has been successfully applied to the sentence-level NMT model~\cite{shaw-etal-2018-self} and other pre-training models~\cite{dai-etal-2019-transformer}.
In this paper, we enhance the document-level seq2seq model to be aware of relative position information in a long input sequence. Note that we integrate relative position into the self-attention module both in the encoder and decoder side.

\subsection{Applying P-Transformer to Multiple NMT Models}

\begin{figure}[t]
\begin{center}
  \includegraphics[width=8.0cm]{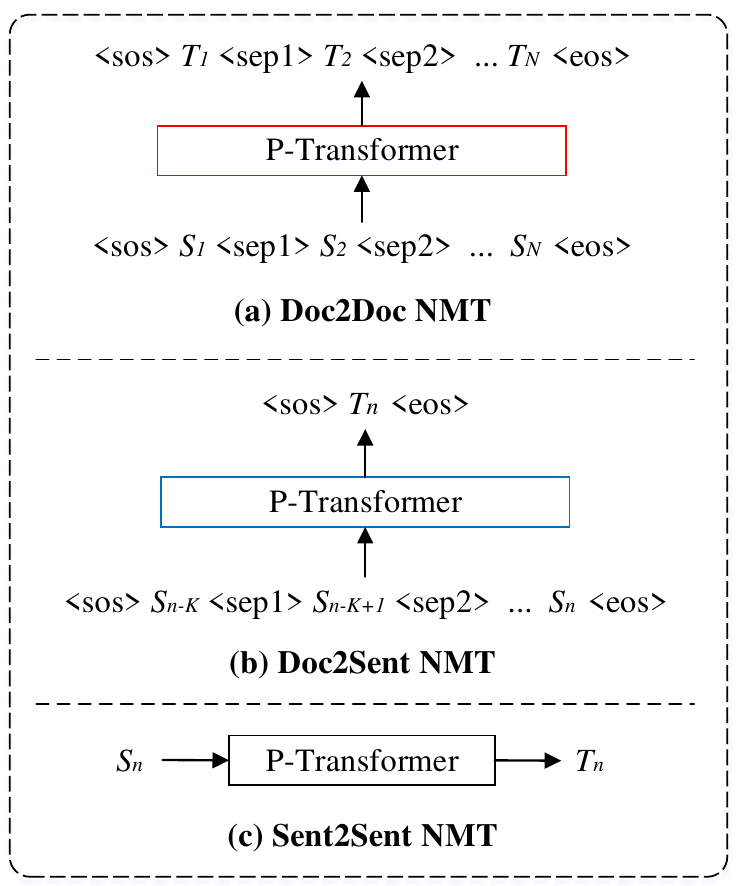}
\end{center}
\caption{
Overview of our proposed P-Transformer-based Doc2Doc, Doc2Sent, and Sent2Sent NMT models.  }
\label{fig:archs}
\end{figure}

We use a parallel document pair {\small $(\mathcal{S}, \mathcal{T}) = \{S_n, T_n\}|_{n=1}^N$} with {\small$N$} sentence pairs to illustrate how we apply the proposed P-Transformer to the following three types of seq2seq-based translation:

\textbf{Doc2Doc Translation.}
For Doc2Doc translation, we concatenate the sentences within a document, both source-side and target-side, together into a long sequence, as shown in Figure~\ref{fig:archs}~(a). We insert \textless sos\textgreater, \textless eos\textgreater, \textless sep$I$\textgreater~$\left(I \in\{ 1, 2, 3,\cdots\}\right)$ into the sequence to denote the start of a document, the end of a document, and the separator between the $I$-th and the $(I+1)$-th sentences, respectively. With the sentence separators \textless sep$I$\textgreater, it is easy to get sentence-to-sentence alignment between a source document and its translation. 

\textbf{Doc2Sent Translation.}
We follow previous work~\cite{zhang-2018-doc,sun-etal-2022-rethinking} to use $K$ previous sentences as context for Doc2Sent translation. As shown in Figure~\ref{fig:archs}~(b), the source side input is a sequence consisting of the $K$ previous sentences, plus the current sentence while the target side output is the translation of the current sentence. Similar to Doc2Doc translation, we insert \textless sos\textgreater, \textless eos\textgreater, \textless sep$I$\textgreater~$\left(I \in\{ 1, 2,\cdots, K\}\right)$ in the source-side sequence. For the target side sequence, we only insert $\textless sos\textgreater$ and $\textless eos\textgreater$.

\textbf{Sent2Sent Translation.}
 As shown in Figure~\ref{fig:archs}~(c) we can also directly apply P-Transformer to Sent2Sent translation.
 
\section{Experimentation}

\subsection{Experimental Setup}
\label{sect:data}
\noindent\textbf{Datasets.} We carry out extensive experiments on various document-level translation datasets covering 7 language pairs, including Chinese-to-English (Zh-En), English-to-German (En-De), French-to-English (Fr-En), Spanish-to-English (Es-En), Russian-to-English (Ru-En), English-to-Russian (En-Ru) and English-to-French (En-Fr).
The details of the datasets are listed as follows:

\begin{itemize}
\item \textbf{Zh-En}: we use the parallel document corpus (PDC) released by \citet{sun-etal-2022-rethinking} for Zh-En translation.\footnote{https://github.com/sunzewei2715/Doc2Doc\_NMT} The dataset contains middle-scale data with documents from different domains, including politics, finance, health, culture, etc. The training, validation, and test sets are including 1.39M~/~2.0M~/~4.9K sentences, respectively.
 
\item \textbf{En-De}: we use three datasets provided by~\citet{maruf:2019-doc}, including TED, News and Europarl.\footnote{https://github.com/sameenmaruf/selective-attn/tree/master/data} The TED dataset is from IWSLT 2017\footnote{https://wit3.fbk.eu/}~\cite{cettolo-2012-wit3} and we use tst2016-2017 as the test set, and other as the validation set. The News dataset is from the News Commentary v11  corpus.\footnote{http://www.casmacat.eu/corpus/news-commentary.html} We use WMT newstest2015 as the validation set and newstest2016 as the test set, respectively. The Europarl dataset is extracted from Europarl v7, and the training, validation, and test sets are obtained by randomly splitting the corpus. 
The training, validation, and test sets of TED are including 0.21M~/~9.0K~/~2.3K sentences, respectively.
The training, validation, and test sets of News are including 0.24M~/~2.0K~/~3.0K sentences, respectively.
The training, validation, and test sets of Europarl are including 1.67M~/~3.6K~/~5.1K sentences, respectively.

\item \textbf{Fr-En, Es-En and Ru-En}: we use News Commentary v14 (News14) from WMT 2019\footnote{https://www.statmt.org/wmt19/translation-task.html} as their training data. The validation sets~/~test sets are newstest2012~/~newstest2013 for Es-En translation, newstest2013~/~newstest2014 for FR-EN translation, newstest2018~/~newstest2019 for Ru-En translation, respectively.
The training, validation, and test sets of Fr-En are including 0.33M~/~3.0K~/~3.0K sentences, respectively.
The training, validation, and test sets of Es-En are including 0.38M~/~3.0K~/~3.0K sentences, respectively.
The training, validation, and test sets of Ru-En are including 0.28M~/~3.0K~/~2.0K sentences, respectively.
\item \textbf{En-Ru and En-Fr}: we choose the widely used large-scale OpenSubtitles~2018 (Subtitles) datasets.\footnote{https://opus.nlpl.eu/OpenSubtitles2018.php} The original En-Ru and En-Fr datasets contain 25.9M~/~41.8M sentences. we split long documents into sub-documents with up to 512 tokens, resulting in 0.48M~/~0.81M sub-documents. For En-Ru and En-Fr translations, we randomly selected 0.36M~/~2K~/~1K and 0.60M~/~2K~/~1K documents as the training set, validation set, and test set from the processed document-level instances, and there is none-overlapping between them.
\end{itemize}

\begin{table}[!t]
\caption{\label{tbl:statis} Statistics on the pre-processed document-level translation data sets. ``S'', ``M'' and ``L'' denote that the corresponding training data is in small-, middle- or large-scaled, respectively. } 
\begin{center}
	\begin{tabular}{ll|c|cc}
		\hline 
		&\multirow{2}*{\bf{Dataset}}&\multirow{2}*{\bf{Type}}  & \bf \#Doc  & \bf Avg \# Token/Doc \\ 
		 &&& (train / valid / test) & (train / valid / test) \\
		\hline
		Zh-En & PDC & M & 119.8K / 255 / 320 & 383 / 282 / 378 \\ \hline
		\multirow{3}*{En-De} & TED & S & 10.9K / 468 / 115 & 451 / 444 / 445 \\
		& News & S & 17.9K / 165 / 249 & 405 / 365 / 338 \\ 
        & Europarl & M & 150.8K / 322 / 458 &  334 / 343 / 342\\
		\hline
		Fr-En & News14 & S & 24.7K / 201 / 274 & 430 / 436 / 352 \\ \hline
		Es-En & News14 & S & 28.0K / 229 / 192 & 424 / 414 / 443 \\ \hline
		Ru-En & News14 & S & 21.4K / 266 / 213 & 411 / 345 / 258 \\ \hline
		En-Ru & Subtitles & L & 360K / 2000 / 1000 & 486 / 487 / 488 \\ \hline
		En-Fr & Subtitles & L & 600K / 2000 / 1000 & 478 / 479 / 478 \\ \hline
	\end{tabular}
\end{center}
\end{table}

\begin{table*}[!t]
\centering
\caption{\label{tbl:main} BLEU scores on Zh-En and En-De translations. \dag/\ddag: significant over Transformer~(sent) baselines at 0.05/0.01.}
\begin{tabular}{c|cl|cc|cccccc}
\hline 
\multirow{3}{*} {\bf Type} & \multirow{3}{*} {\#} & \multirow{3}{*} {\bf Model} & \multicolumn{2}{c|}{\bf Zh-En} & \multicolumn{6}{c}{\bf En-De} \\ 
& & & \multicolumn{2}{c|}{\bf PDC} & \multicolumn{2}{c}{\bf TED} & \multicolumn{2}{c}{\bf News} & \multicolumn{2}{c}{\bf Europarl} \\ 
\cline{4-11}
& & & s-BLEU & d-BLEU & s-BLEU & d-BLEU & s-BLEU &d-BLEU & s-BLEU & d-BLEU \\ \hline
\multirow{6}{*}{Sent2Sent} & \multicolumn{10}{c}{Existing Models} \\
\cline{2-11}
& 1 & Transformer \cite{bao-etal-2021-g} & - & - & 24.82 & - & 25.19 & - & 31.37 & - \\
& 2 & Transformer \cite{sun-etal-2022-rethinking} & - & - & 25.19 & - & 24.98 & - & 31.70 &  \\ 
\cline{2-11}
& \multicolumn{10}{c}{Our Models} \\
\cline{2-11}
& 3 & Transformer (sent) & 25.80 & - & 24.83 & - & 25.13 & - & 31.55 & - \\ 
& 4 & P-Transformer (sent) & \textbf{26.13}\dag & - & \textbf{25.26}\dag & - & \textbf{25.57}\dag & - & \textbf{31.85}\dag & \\
\hline
\hline
\multirow{9}{*}{Doc2Sent} & \multicolumn{10}{c}{Existing Models} \\
\cline{2-11}
& 5 & DocT \cite{zhang-2018-doc} & - & - & 24.00 & & 23.08 & & 29.32 &- \\ 
& 6 & HAN \cite{miculicich:2018-doc} & - & - &  24.58 & - & 25.03 & -& 28.60 & -\\ 
& 7 & SAN \cite{maruf:2019-doc} & - & - &24.42 & -& 24.48 & -& 29.75 & -\\
& 8 & QCN \cite{yang_etal_emnlp-ijcnlp_2019} & - & - & 25.19 & - & 22.37 & - & 29.82 & - \\
& 9 & Flat-Tran. \cite{ma:2020-simple} & - & - & 24.87 & - & 23.55 & - & 30.09 & - \\
& 10 & MCN \cite{zheng:2020-ijcai} & - & - & 25.10 & - & 24.91 &- & 30.40 & -\\
& 11 & MR Doc2Sent \cite{sun-etal-2022-rethinking} & - & - &25.24 & 29.20 & 25.00 & 26.70 & 32.11 & 34.18 \\
\cline{2-11}
& \multicolumn{10}{c}{Our Models} \\
\cline{2-11}
& 12 & P-Transformer (ctx) & 25.50 & - & 25.47\ddag &  & 25.15 & & 31.80\dag & \\
& 13 & P-Transformer (ctx + sent) & \textbf{26.43}\ddag & - & \textbf{25.73}\ddag & - & \textbf{25.71}\ddag & - & \textbf{32.15}\ddag & \\
\hline
\hline
\multirow{9}{*}{Doc2Doc} & \multicolumn{10}{c}{Existing Models} \\
\cline{2-11}
& 14 & G-Transformer \cite{bao-etal-2021-g} & - & - & 23.53 & 25.84 & 23.55 & 25.23 & 32.18 & 33.87 \\
& 15 & G-Transformer (fine-tuned) \cite{bao-etal-2021-g} & - & - & 25.12 & 27.17 & 25.52 & 27.11 & 32.39 & 34.08 \\
& 16 & SR Doc2Doc \cite{sun-etal-2022-rethinking} & - & 24.33 & - & 4.70 & - & 21.18 & - & 34.16 \\
& 17 & MR Doc2Doc \cite{sun-etal-2022-rethinking} & - & 27.80 & - & 29.27 & - & 26.71 & - & 34.48 \\
\cline{2-11}
& \multicolumn{10}{c}{Our Models} \\
\cline{2-11}
& 18 & Transformer (doc) & 25.38 & 27.39 & - & 0.76 & - & 0.60& 31.54 & 33.21 \\
& 19 & P-Transformer (doc) & 26.53\ddag & 28.53 & 24.91 & 27.09 & 24.92 & 27.03 & 32.37\ddag & 34.14 \\
& 20 & P-Transformer (doc + sent) & \textbf{27.14}\ddag & \textbf{29.26} & \textbf{25.67}\ddag & 27.94 & \textbf{25.93}\ddag & \textbf{27.67} & \textbf{32.62}\ddag & \textbf{34.49} \\
\hline
\end{tabular}
\begin{tablenotes}
\item \textit{Note}: (sent), (ctx), (doc) indicate the model is trained on Sent2Sent, Doc2Sent and Doc2Doc training instances, respectively. (ctx/doc + sent) indicate the model is trained on Doc2Sent/Doc2Doc and Sent2Sent training instances. 
\end{tablenotes}
\label{tab:my_label}
\end{table*}

In pre-processing, we tokenize all sentences by Moses toolkit\footnote{https://github.com/marian-nmt/moses-scripts} while the Chinese sentences are segmented by the widely used Jieba toolkit.\footnote{\url{https://github.com/fxsjy/jieba}} Then, the source and target sentences are segmented into sub-words by a joint BPE model~\cite{sennrich_etal:16} with 32K merged operations. Note that the pre-processing is identical to the related work of G-Transformer~\cite{bao-etal-2021-g} and MR Doc2Doc~\cite{sun-etal-2022-rethinking}. 
Finally, following G-Transformer~\cite{bao-etal-2021-g}, we split long documents into sub-documents with up to 512 tokens in Doc2Doc NMT experiments. Table~\ref{tbl:statis} shows the detailed statistics of datasets after pre-processing.

\noindent\textbf{Model Setting.} We use \textit{Fairseq}~\cite{ott_etal:2018} as the implementation of Transformer models. We follow the standard Transformer base model setting~\cite{vaswani_etal:17}, in which we use 6 layers, 8 heads, 512 dimension outputs, and 2048 dimension hidden states. Note that the position information we add to attention is always sinusoidal PEs.
In all experiments, we use the learning rate decay policy proposed by~\citet{vaswani_etal:17} (warm-up step 4K) with label smoothing of 0.1, and the dropout is 0.3. We share bilingual vocabulary to reduce the computation cost.
In inference, we choose the best checkpoint on the validation set to evaluate the translation performance. We set the beam size as 5, and the three parameters to control the generation length lenpen~/~max-len-a~/~max-len-b as 1.0~/~1.1~/~7, respectively. 
We run all experiments 3 times with 3 different random seeds on the small- and middle-scaled Zh-En, En-De, En-Fr, En-Es, and En-Ru tasks and reported averaged BLEU scores. Since the big load of computation, the experiments on large-size datasets only run once.

\noindent\textbf{Training.} All the models are trained on a single Tesla V100 GPU. We set the gradient update frequency~/~token size as 4~/~8192 for the Transformer model on middle- and large-scaled PDC, Europarl, and Subtitles datasets, 4~/~4096 for other small datasets. Finally, we stop training by using an early stopping strategy on the validation set for small- and middle-scaled datasets. For large Subtitles datasets, we train NMT models with 20 epochs on both En-Ru and En-Fr translations. As fine-tuning the sentence-level NMT model benefits Doc2Doc model~\cite{bao-etal-2021-g,liu-2020-mbart}, we also combine both Doc2Sent (or Doc2Doc) training instances and Sent2Sent training instances when training P-Transformer-based Doc2Sent (or Doc2Doc) models.

\noindent\textbf{Evaluation.} Following previous related work, we report case-sensitive document-level BLEU~\cite{liu-2020-mbart} (d-BLEU)  for the document-level NMT systems, which is computed by matching n-grams in the whole document after removing the special tokens of \textless sos\textgreater, \textless eos\textgreater, and \textless sep$I$\textgreater. 
Thanks to the proposed sentence separators, we can obtain the sentence-to-sentence alignments between a source document and its translation. So we also report the conventional case-sensitive sentence-level BLEU (s-BLEU).\footnote{As shown in Table\ref{tbl:doc-preds}, very few source sentences have no alignment in the translation. For these source sentences, we set their translation as empty when calculating s-BLEU.}

\subsection{Main Results of Sent2Sent Translation and Doc2Sent Translation}

Table~\ref{tbl:main} (Sent2Sent) shows the BLEU scores (s-BLEU) of the Sent2Sent NMT models.
The performance of our Transformer baseline (\#3) is comparable with that of related studies~(\#1 and \#2).
Moreover, our proposed P-Transformer (\#4) improves the translation quality of Sent2Sent NMT on all datasets, which suggests that even though the position information is well preserved at the top layer of the Transformer encoder, it also benefits from our proposed position-aware attention. 

Table~\ref{tbl:main} (Doc2Sent) presents the experimental results of Doc2Sent models.
Note that in the Doc2Sent instances, the source side input is a sequence consisting of the 4 previous sentences, plus one current sentence.
From the table, we observe that with the help of contextual sentences, the translation performance of P-Transformer~(ctx) (\#12) is improved on En-De translations. This shows that a larger source context benefits translation performance. 
By including sentence-level training data, P-Transformer (ctx + sent) (\#13) further improves the translation quality on both Zh-En and En-De translations.
Compared with related studies in Doc2Sent, our model achieves the best performance.

\subsection{Main Results of Doc2Doc NMT}

\noindent\textbf{Results of Zh-En and En-De translations.} As shown in Table~\ref{tbl:main} (Doc2Doc), it fails to train a Doc2Doc model with vanilla Transformer (\#18) on small-scaled datasets (e.g., TED and News), while it can be successfully trained on the middle-scaled PDC and Europarl datasets. This is consistent with the findings in G-Transformer\cite{bao-etal-2021-g}. Different from the vanilla Transformer, the proposed P-Transformer~(\#19) can be successfully trained on all datasets. Compared with Sent2Sent~(\#3) baseline systems, P-Transformer (\#19) trained directly on document-level data achieves comparable performance on small-scaled TED and News datasets, or significantly better performance on middle-size PDC and Europarl datasets, which shows good capability of P-Transformer in document-level context modeling.
Moreover, P-Transformer (\#19) outperforms both G-Transformer and SR Doc2Doc (\#14 and \#16) when direct training on document-level data, especially on small-scale datasets.
Finally, compared with G-Transformer~(\#15) and MR Doc2Doc~(\#17), P-Transformer with additional sentence-level training data (\#20) achieves a new state-of-the-art on PDC, TED, News, and Europarl.
Note that our model (\#20) is lower than both MR Doc2sent (\#11) and MR Doc2Doc (\#17) models on the TED data set measured by d-BLEU, but it outperforms MR Doc2Sent~(\#11) in s-BLEU, while the two systems~(\#11 and \#17) get similar performance in d-BLEU.

\begin{table*}[!ht]
\caption{\label{tbl:main-others} BLEU scores of document-level NMT models on Fr-En, Es-En, and Ru-En translation tasks. \dag/\ddag: significant over Transformer~(sent) baselines at 0.05/0.01. }
\begin{center}
\begin{tabular}{l|l|cc|cc|cc}
\hline 
\multirow{2}{*} {Type} & \multirow{2}{*} {Model} & \multicolumn{2}{c|}{Fr-En} & \multicolumn{2}{c|}{Es-En} & \multicolumn{2}{c}{Ru-En} \\
\cline{3-8}
& & s-BLEU & d-BLEU & s-BLEU & d-BLEU & s-BLEU &d-BLEU \\ 
\hline
Sent2Sent & Transformer (sent) & 28.11 & - & 28.21 & - & 22.03 & - \\
\hline
\multirow{5}{*}{Doc2Doc} & SR Doc2Doc \cite{sun-etal-2022-rethinking}& - & 23.86 & - & 26.79 & - & 16.47 \\ 
& MR Doc2Doc \cite{sun-etal-2022-rethinking}& - & 28.85 & - & 29.37 & - & 23.98 \\ 
\cline{2-8}
& Transformer (doc) & - & 0.79 & - & 0.73 & - & 0.06 \\
& P-Transformer~(doc) & 28.29 & 29.94 & 28.10 & 30.65 & 21.56 & 23.96 \\
& P-Transformer~(doc + sent) & \textbf{29.16}\ddag & \textbf{30.79} & \textbf{28.98}\ddag & \textbf{31.26} & \textbf{22.72}\ddag & \textbf{24.39} \\
\hline
\end{tabular}
\end{center}
\end{table*}

\noindent\textbf{Results of Fr-En, Es-En and Ru-En translations.} 
Table~\ref{tbl:main-others} shows the translation performance on Fr-En, Es-En, and Ru-En translations. It is unsurprising to observe that vanilla Transformer fails to be successfully trained on the three small-scale datasets. When being trained on document-level data, P-Transformer (doc) achieves similar performance on Fr-En and Es-En translation compared to sentence-level Transformer while it achieves better or comparable performance than MR Doc2Doc \cite{sun-etal-2022-rethinking}. By including sentence-level training data, P-Transformer~(doc~+~sent) achieves the best performance over the three translation tasks. 

\begin{table}[!t]
\caption{\label{tbl:main-large} BLEU scores of document-level NMT models on large-scale En-Ru and En-Fr translations, respectively. \dag/\ddag: significant over Transformer~(sent) baselines at 0.05/0.01. }
\begin{center}
\begin{tabular}{l|cc|cc}
\hline 
\multirow{2}{*} {Models} & \multicolumn{2}{c|}{En-Ru} & \multicolumn{2}{c}{En-Fr} \\
\cline{2-5}
& s-BLEU & d-BLEU & s-BLEU & d-BLEU \\ 
\hline
Transformer~(sent) & 25.80 & - & 33.89 & - \\
Transformer~(doc) & 24.32 & 28.16 & 32.82 & 36.70 \\  \hline
P-Transformer~(doc) & 24.94 & 28.81 & 34.24\ddag & 38.07 \\
\scriptsize{P-Transformer~(doc + sent)} & \textbf{25.98}\dag & \textbf{29.83} & \textbf{34.36}\ddag & \textbf{38.19} \\
\hline
\end{tabular}
\end{center}
\end{table}

\noindent\textbf{Results of En-Ru and En-Fr translations.}
Table~\ref{tbl:main-large} shows the translation performance on large-scale En-Ru and En-Fr translation tasks. From it, we observe that our proposed P-Transformer~(doc) significantly outperforms Transformer~(doc) in terms of both s-BLEU and d-BLEU.
For example, P-Transformer (doc) achieves gains of 1.42~/~1.37 on s-BLEU / d-BLEU over Transformer (doc) in En-Fr translation. 
Similarly, by taking advantage of sentence-level training data, P-Transformer (doc + sent) achieves the best performance, with further improved translation performance.
From the above experimental results, we see that the proposed P-Transformer has a strong generalization ability on context modeling over different language pairs and dataset scales. 

\subsection{Parameters and Training Speed}

\begin{table}[!th]
\centering
\caption{Parameter (in Millions) and training speed~(second / epoch).\label{tbl:para}}
\begin{tabular}{cl|c|c}
\hline
\bf \# & \bf Model & \bf \#Param. & \bf Speed \\
\hline
1 & Transformer (sent) & 59.38M & 460s \\
\hline
2 & P-Transformer (sent) & 59.45M & 473s \\
3 & ~~-rel-self & 59.38M & 462s \\
\hline
4 & P-Transformer (doc) & 59.45M & 647s \\
5 & ~~-rel-self & 59.38M & 562s \\
\hline
\end{tabular}
\begin{tablenotes}
\item \textit{Note:} ``-rel-self'' denotes removing the relative position encoding in self-attention.
\end{tablenotes}
\end{table}

We use models on the En-De TED translation task as an example to analyze model parameters and training speed, as shown in Table~\ref{tbl:para}. Comparing \#1 and \#3, we observe that our proposed position-aware self-attention and cross-attention bring no additional parameters and negligible computation cost. Moreover, the relative position for self-attention only introduces a matrix of relative position embeddings with 1025$\times$64=65,600 additional parameters and requires slightly more computation cost (as shown in \#2 VS. \#3, and \#4 VS. \#5).
It is well known that the computational complexity of self-attention is $I^2 \times D$, where $I$ is the sequence length, and $D$ is the hidden state size. Compared with the sentence-level baseline model, the training time of an epoch for P-Transformer~(doc) increases by about 40\% (as shown in \#1 VS. \#4).

\section{Analysis}

In this section, we take Doc2Doc NMT models as representatives to discuss the effectiveness of P-Transformer, which unless otherwise specified, is trained on document-level instances, i.e., P-Transformer (doc).

\subsection{Analysis on Different Components}
We explore the contributions of position-aware self-attention, cross-attention, and relative position encoding for Doc2Doc P-Transformer. Table~\ref{tbl:comp} shows the performance in d-BLEU on the three En-De translation tasks.

On the one hand, P-Transformer fails on TED and News when we disable position-aware attention in the cross-attention module (-cross-attn). On the other hand, P-Transformer still can be properly trained when we disable position-aware attention in the three self-attention modules (-rel-self, -self-src, -self-tgt). The performance trend suggests that the cross-attention module is the key for Doc2Doc NMT while the two self-attention modules can also benefit from position information.

\begin{table}[!ht]
\caption{\label{tbl:comp} Performance (d-BLEU) on the three En-De translation tasks.}
\begin{center}
\begin{tabular}{l|ccc|c}
\hline \bf Model & \bf TED & \bf News & \bf \small{Europarl} & \bf Drop \\ 
\hline
P-Trans. (doc) & \textbf{27.09} &  \textbf{27.03} & \textbf{34.14} & - \\ \hline
-cross-attn & 0.70 & 0.83 & 33.67 & -17.53 \\ 
-self-src, -self-tgt& 24.05 & 26.05 & 33.78 & -1.34 \\ \hline
-self-src & 25.67 & 25.12 & 33.90 & -1.11 \\
-self-tgt &  26.62 & 26.12 & 33.93 & -0.46 \\
\hline
-rel-self & 26.75 & 26.61 & 34.01 & -0.25 \\ \hline
\end{tabular}
\begin{tablenotes}
\item \textit{Note:} ``-cross-attn'', ``-self-src'', and ``-self-tgt'' denote removing the absolute position embeddings in the cross-attention, source-side self-attention, and target-side self-attention, respectively while ``-rel-self'' denotes removing the relative position embeddings in self-attention.
\end{tablenotes}
\end{center}
\end{table}

\subsection{Analysis on Sentence Separator in Doc2Doc Translation} 
In this section, we first analyze the sentence-level coverage issue in Doc2Doc translation. Then we compare the performance of using different sentence separator methods. 

\begin{table}[!t]
\caption{\label{tbl:doc-preds} Statistics on the test sets, regarding documents without sentence-level coverage issue in Doc2Doc translation.}
\begin{center}
	\begin{tabular}{l|ccc}
		\hline \bf Dataset & \bf Correct & \bf Total & \bf Percentage \\ 
		\hline 
         Zh-En (PDC) & 316 & 317 & 99.7\% \\
         \hline
         En-De (TED) & 114 & 115 & 99.1\% \\
         En-De (News) & 245 & 249 & 98.4\% \\
         En-De (Europarl) & 458 & 458 & 100\% \\
         \hline
         Fr-En (News14) & 273 & 274 & 99.6\% \\
         \hline
         Es-En (News14) & 188 & 192 & 97.9\% \\
         \hline
         Ru-En (News14) & 209 & 213 & 98.1\% \\
         \hline
         En-Ru (Subtitles) & 991 & 1000 & 99.1\% \\
         \hline
         En-Fr (Subtitles) & 985 & 1000 & 98.5\% \\
		\hline
		\textbf{Overall} & \textbf{3779} & \textbf{3818} & \textbf{99.0\%}\\
		\hline
	\end{tabular}
\end{center}
\end{table}

In Doc2Doc translation, we insert the separator \textless sep$I$\textgreater ~between the $I$-th and the $I+1$-th sentences in both the source and target sides. Therefore, we recover sentence-level translation from these separators. We say a document having no sentence-level coverage issues if we can perfectly recover sentence-level translation for all its source sentences. From Table~\ref{tbl:doc-preds}, we find that 97.9\% to 100\% documents are properly translated without sentence-level coverage issues. 
For those few documents having sentence-level coverage issues, further analysis reveals that most of them miss one or two sentences at the end of translation. 

\begin{table}[!ht]
\caption{\label{tbl:seps} Performance (s-BLEU / d-BLEU) comparison of using different sentence separator methods.}
\begin{center}
\begin{tabular}{l|c|ccc}
\hline \bf Model & \bf Sep. & \bf TED & \bf News & \bf Europarl \\ 
\hline
P-Trans. &\textless sep$I$\textgreater& \textbf{24.91} / 27.09  & \textbf{24.92} / \textbf{27.03} & \textbf{32.37}/ \textbf{34.14} \\ \hline
P-Trans. &\textless sep\textgreater& 24.37 / \textbf{27.25}  & 23.23 / 26.71  & 32.06 / 33.96  \\
\hline
\end{tabular}
\begin{tablenotes}
\item \textit{Note:} Here P-Trans. indicates P-Transformer (doc) model.
\end{tablenotes}
\end{center}
\end{table}

Instead of using index-aware sentence separators, i.e., \textless sep$I$\textgreater, we compare its performance with another method that uses a uniform sentence separator \textless sep\textgreater. 
As shown in Table~\ref{tbl:seps}, we find that overall the uniform sentence separator method under-performs the index-aware sentence separator method. Moreover, further analysis of the test sets reveals that the uniform sentence separator suffers more from the sentence-level coverage issue. 

\subsection{One Model Learns both Sent2Sent and Doc2Doc Translation}

As mentioned, our P-Transformer (doc + sent) in Table~\ref{tbl:main}~(Doc2Doc) is trained on both sentence-level and document-level sequences. Therefore, the trained model could translate both sentences and documents. Table~\ref{tbl:main-sent} compares the performance of s-BLEU when the input unit in inference is a sentence or document. 
\begin{table}[!ht]
\caption{\label{tbl:main-sent} Performance comparison on s-BLEU score of Sent2Sent and Doc2Doc NMT models.}
\begin{center}
\begin{tabular}{l|c|cccc}
\hline \bf Model & \bf Input & \bf PDC &\bf TED & \bf News & \bf Europarl \\ 
\hline 
Trans. (sent) & sent & 25.80 & 24.83 & 25.13 & 31.55 \\
P-Trans. (sent) & sent & 26.13 & 25.26 & 25.57 & 31.85 \\ 
\hline
P-Trans. (doc + sent) & sent & 26.40 & 25.21 & 25.37 & 31.99 \\
P-Trans. (doc + sent) & doc & \textbf{27.14} & \textbf{25.67} &  \textbf{25.93} & \textbf{32.62} \\
\hline
\end{tabular}
\end{center}
\end{table}

Table~\ref{tbl:main-sent} shows that no matter whether the input unit is a sentence or document, P-Transformer (doc + sent) achieves better performance than sentence-level Transformer baseline. This shows that our proposed document-level seq2seq model is beneficial for both document- and sentence-level NMT. This result demonstrates that a single P-Transformer (doc + sent) model can be used for both Sent2Sent and Doc2Doc translation tasks. Moreover, it shows Doc2Doc translation benefits from using document-level context. 

\subsection{Analysis on Discourse Phenomena}

To verify whether Doc2Doc NMT truly learns the useful contextual information to improve discourse coherence, we use the linguistic feature test set provided by~\citet{sun-etal-2022-rethinking} to evaluate different discourse phenomena, including tense consistency~(TC), conjunction presence~(CP), and pronoun translation~(PT). TCP is an overall score calculated as the geometric mean of TC, CP, and PT, which has a strong correlation with human evaluation~\cite{sun-etal-2022-rethinking}.

\begin{table}[!ht]
\caption{\label{tbl:dis-zhen} Discourse phenomena evaluation on the Zh-En test set. }
\begin{center}
\begin{tabular}{l|c|ccc|c}
\hline 
\bf Model & \bf s-BLEU & \bf TC & \bf CP & \bf PT & \bf TCP \\ 
\hline
Transformer (sent) & 25.80 & 57.3 & 34.7 & 61.6 & 49.6 \\
Transformer (doc) & 25.38 & 55.7 & 31.9 & 60.8 & 47.6 \\ \hline
P-Transformer (doc) & 26.53 & 57.7 & \textbf{37.4} & 62.2 & \textbf{51.2} \\
P-Transformer (doc + sent) & \textbf{27.14} & \textbf{58.2} & 33.6 & \textbf{64.9} & 50.3 \\
\hline
\end{tabular}
\end{center}
\end{table}

As shown in Table~\ref{tbl:dis-zhen}, compared with Transformer~(doc) and Transformer~(sent) systems, our proposed P-Transformer~(doc) improves not only the translation performance in BLEU but also the performance of discourse phenomena.
By including sentence-level data, our P-Transformer~(doc + sent) further improves TC and PT, but decreases both CP and overall TCP. This suggests that though sentence-level data improves the document-level NMT quality in BLEU, it has an uncertain effect on various discourse phenomena, e.g., improving performance in TC and PT while decreasing performance in CP and overall TCP.

\subsection{Analysis on Transformer Encoder Depth}
As our probing task result suggests that position information gradually vanished from the bottom encoder layers to the up layers in vanilla Transformer. Therefore, decreasing its encoder depth might make the training successful, especially on small datasets. 

\begin{figure}[ht]
\setlength{\abovecaptionskip}{0pt}
\begin{center}
\pgfplotsset{height=4.4cm,width=7.0cm,compat=1.13}
\begin{tikzpicture}
\tikzset{every node}=[font=\scriptsize]
\begin{axis}[
	ylabel=BLEU,
	enlargelimits=0.02,
	tick align=outside,
	legend style={at={(0.5,-0.17)},
	anchor=north,legend columns=-1},
	ybar interval=0.6,
	xticklabels={\tiny{Transformer-4},\tiny{Transformer-5},\tiny{Transformer-6},\tiny{P-Transformer},8},
	xtick={1,2,3,4,5},
	ytick distance=5,
]
\addplot+ [color=black,fill=black,pattern=dots,pattern color=black] coordinates{(1,24.97) (2,25.39) (3,0.41) (4,27.09) (5,0)};
\addplot+ [color=blue,fill=blue,pattern color=blue,pattern=vertical lines] coordinates{(1,24.55) (2,25.16) (3,0.52) (4,27.03) (5,0)};
\addplot+ [color=blue,fill=blue,pattern color=blue,pattern=north east lines] coordinates{(1,32.85) (2,33.14) (3,33.21) (4,34.14) (5,0)};
\legend{TED,News, Europarl}
\end{axis}
\end{tikzpicture}
\caption{BLUE scores when using different encoder depths for Doc2Doc NMT on the three En-De translation tasks.  ``Transformer-$X$'' represents vanilla Transformer with $X$ encoder layers. }\label{fig:ana-depth}
\end{center}
\end{figure}
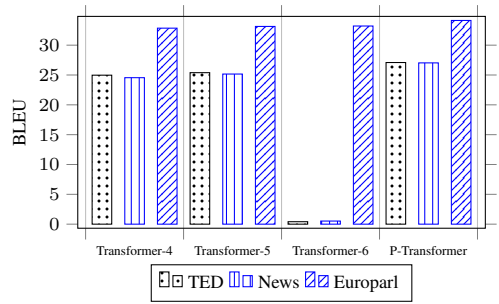

As shown in Figure~\ref{fig:ana-depth}, vanilla Transformer models with encoder depth of either 4 or 5 achieve good performance on all datasets. However, decreasing encoder depth would weaken the capability of capturing useful information from input sequences, thus lowering the translation performance. 

\subsection{Analysis on Input Length}
Next, we investigate the effect of the input length. To this end, we split long documents of TED into sub-documents with up to $I$ tokens, where $I\in\{64, 128, 256, \dots\}$. 
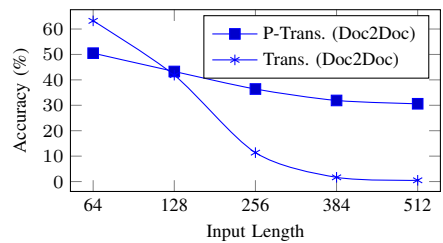
\begin{figure}[ht]
	\setlength{\abovecaptionskip}{0pt}
	\begin{center}
		\pgfplotsset{compat=1.13}
		\begin{tikzpicture}
		\tikzset{every node}=[font=\scriptsize]
		\begin{axis}
		[height=4.0cm,width=6.5cm,enlargelimits=0.07, tick align=inside, legend style={cells={anchor=west},legend pos=north east}, xticklabels={64,128,256,384,512},
		xtick={1,2,3,4,5},
		ylabel={Accuracy (\%)},
		xlabel={Input Length},
		ytick distance=10,
		]	
		\addplot+[smooth,color=blue,mark color=blue,mark=square*] coordinates
		{(1,50.5) (2,43.3) (3,36.4) (4,31.9) (5,30.6)};
		\addlegendentry{\scriptsize{P-Trans. (Doc2Doc)}}
		\addplot+[smooth,color=blue,mark color=blue,mark=asterisk] coordinates
		{(1,63.21) (2,41.86) (3,11.38) (4,1.66) (5,0.49)};
		\addlegendentry{\scriptsize{Trans. (Doc2Doc)}}
		\end{axis}
		\end{tikzpicture}
		\caption{The accuracy of position information probing task with respect to different input lengths.}\label{fig:prob-sys}
	\end{center}
\end{figure}

Figure~\ref{fig:prob-sys} compares the accuracy of the absolute position probing task of the top encoder layer. It shows that 1) for Transformer, the accuracy significantly decreases when the input length exceeds 128 and the position information has almost vanished when the length is over 256; 2) for P-Transformer, the encoder still captures certain position information even when the input length increases to 512. 

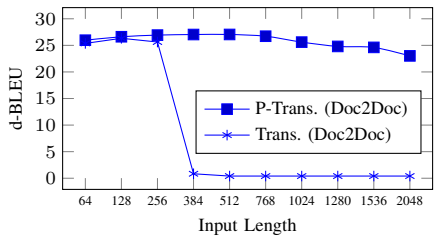
\begin{figure}[ht]
	\setlength{\abovecaptionskip}{0pt}
	\begin{center}
		\pgfplotsset{compat=1.13}
		\begin{tikzpicture}
		\tikzset{every node}=[font=\scriptsize]
		\begin{axis}
		[height=4.0cm,width=6.5cm,enlargelimits=0.07, tick align=inside, legend style={cells={anchor=west},legend pos=south east, yshift=0.4cm}, xticklabels={\tiny{64},\tiny{128},\tiny{256},\tiny{384},\tiny{512},\tiny{768},\tiny{1024},\tiny{1280},\tiny{1536},\tiny{2048}},
		xtick={1,2,3,4,5,6,7,8,9,10},
		ylabel={d-BLEU},
		xlabel={Input Length},
		ymax=30,
		ymin=0,
		ytick distance=5,
		]	
		\addplot+[smooth,color=blue,mark color=blue,mark=square*] coordinates
		{(1,25.97) (2,26.62) (3,26.93) (4,27.04) (5,27.05) (6,26.72) (7,25.61) (8,24.80) (9,24.62) (10,23.02) };
		\addlegendentry{\scriptsize{P-Trans. (Doc2Doc)}}
		
		\addplot+[sharp plot,color=blue,mark color=blue,mark=asterisk] coordinates
		{(1,25.37) (2,26.33) (3,25.63) (4,0.84) (5,0.41) (6,0.41) (7,0.41) (8,0.41) (9,0.41) (10,0.41) };
		\addlegendentry{\scriptsize{Trans. (Doc2Doc)}}
		
		\end{axis}
		\end{tikzpicture}
		\caption{d-BLEU scores with respective to the different input lengths.}\label{fig:length}
	\end{center}
\end{figure}

Figure~\ref{fig:length} compares the translation performance on the TED dataset. It shows that the vanilla Transformer benefits from increasing the input length from 64 to 128 and its performance starts to decrease when increasing the length to 256. Moreover, translation is completely failed when the input length is up to 384 or longer. On the other hand, P-Transformer is less sensitive to the input length, as its translation performance is getting higher when increasing the input length from 64 to 768. Even when the input length is as long as 2048, it still achieves a 23 d-BLEU score.

\begin{table}[!ht]
\caption{\label{tbl:pdc-leng} Performance of P-Transformer on Zh-En PDC dataset.}
\begin{center}
\begin{tabular}{l|cc}
\hline
\bf Input Length & \bf s-BLEU & \bf d-BLEU\\
\hline
512 & 26.53 & 28.53 \\
1024 & \textbf{26.74} & \textbf{29.17} \\
2018 & 25.91 & 28.58\\
\hline
\end{tabular}
\end{center}
\end{table}

To further verify the performance of P-Transformer on long document translation, we split documents of Zh-En PDC dataset into sub-documents with up to $J$ tokens, where $J\in\{512, 1024, 2048\}$.
From Table~\ref{tbl:pdc-leng}, we find that in the middle-scale PDC dataset, the translation performance is stable and robust over different lengths, in which the document of 1024 length achieves the best performance.

\subsection{Analysis on Training Data Scale}

To test the performance of the P-Transformer on small training data, we train models on training datasets with different data sizes by randomly selecting 1K $\sim$ 10K sub-documents from En-De TED translation. 

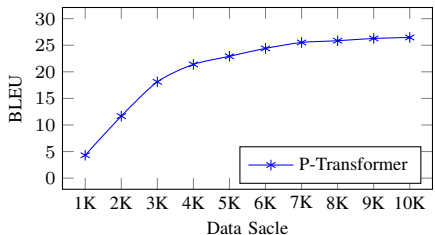
\begin{figure}[ht]
	\setlength{\abovecaptionskip}{0pt}
	\begin{center}
		\pgfplotsset{compat=1.13}
		\begin{tikzpicture}
		\tikzset{every node}=[font=\scriptsize]
		\begin{axis}
		[height=4.0cm,width=6.5cm,enlargelimits=0.07, tick align=inside,legend style={cells={anchor=west},legend pos=south east}, xticklabels={1K,2K,3K,4K,5K,6K,7K,8K,9K,10K},
		xtick={1,2,3,4,5,6,7,8,9,10},
		ylabel={BLEU},
		xlabel={Data Sacle},
		ymax=30.0,
		ymin=0.0,
		ytick distance=5,]	
		\addplot+[smooth,color=blue,mark color=blue,mark=asterisk] coordinates
        {(1,4.30) (2,11.67) (3,18.11) (4,21.39) (5,22.92) (6,24.41) (7,25.52) (8,25.85) (9,26.28) (10,26.46) };
		\addlegendentry{\scriptsize{P-Transformer}}
		\end{axis}
		\end{tikzpicture}
		\caption{BLEU scores of P-Transformer on various data scales.}\label{fig:scale}
	\end{center}
\end{figure}

As shown in Figure~\ref{fig:scale}, P-Transformer has a smooth curve of performance when increasing the size of training data from 1K to 10K. It suggests that P-Transformer can be successfully trained for document-level NMT even on extremely small datasets. 

\section{Related Work}

In this section, we discuss related work from three perspectives: Doc2Sent NMT models, Doc2Doc NMT models, and position encoding for Transformer.

\subsection{Doc2Sent NMT Models} Previous studies have proposed various Doc2Sent NMT models and have achieved great success by generating target-side sentences independently. These studies could be further roughly categorized into two groups. Studies of the first group~\cite{tiedemann:2017,jean_etal_arxiv_2017_larger_context,wang_etal_emnlp_2017,bawden_etal_naacl_2018,zhang-2018-doc,voita_etal_acl_2018,yang_etal_emnlp-ijcnlp_2019,ma:2020-simple,wong-etal-acl-2020-contextual,yun-etal-aaai-2020-improving,li_etal_acl_2020_multiencoder,kang_etal_emnlp_2020,zhang-etal-naacl-2021-multi} consider partial source-side context, such as a few previous sentences. These models are strictly sentence-level as document-level context could be viewed as extra input. Some of those studies~\cite{tiedemann:2017,jean_etal_arxiv_2017_larger_context,wang_etal_emnlp_2017,zhang-2018-doc,yang_etal_emnlp-ijcnlp_2019,ma:2020-simple,yun-etal-aaai-2020-improving,li_etal_acl_2020_multiencoder,kang_etal_emnlp_2020,zhang-etal-naacl-2021-multi} aim to extract general useful information from context to help for translating current sentences while some \cite{bawden_etal_naacl_2018,voita_etal_acl_2018,wong-etal-acl-2020-contextual} aim to resolve discourse phenomena. Studies of the second group~\cite{maruf-2018-doc,maruf:2019-doc,tan-etal-2019-hierarchical,tan-etal-2021-coupling,lyu_et_al_emnlp:21,xu-etal-emnlp-2021-document-graph} extend the source-side context from the local context of a few sentences into larger context within the whole document. These models usually take documents as input units and extract useful information from the global context to help translate current sentences. 
However, one obvious disadvantage of these Doc2Sent models is that it is difficult to properly use the target-side context, which is potentially useful for translation.

\subsection{Doc2Doc NMT Models}
Much effort has also been devoted to Doc2Doc translation. 
Concatenating multiple sentences into a unit is the preliminary exploration of Doc2Doc NMT~\cite{tiedemann:2017,agrawal:2018,ma:2020-simple,zhang-etal-2020-long}. However, these approaches are limited within a short context of up to $4 \sim 6$ sentences. Recent studies successfully train vanilla Transformer models for Doc2Doc translation by taking advantage of either large augmented datasets~\cite{junczys:2019,lupo-2022-concate,sun-etal-2022-rethinking} or pre-trained models~\cite{liu-2020-mbart}. Alternatively, \citet{bao-etal-2021-g} find that the training failure is not caused by over-fitting, but by sticking around local minima. Consequently, they propose G-Transformer by introducing local bias to attention to constrain a target sentence to attend to its corresponding source sentence. G-Transformer is the first Doc2Doc NMT model which can be trained from scratch even on a small dataset. However, this approach requires sentence-to-sentence alignment between the source and target sides, which limits its generality for other seq2seq document-level NLP tasks, like text summarization. Following this research line, we find that the failure of training Doc2Doc NMT models from scratch is due to the vanishing of position information at the encoder output. To this end, we propose a position-aware Transformer to enhance both the absolute and relative position information in attention modules. Our position-aware self-attention and cross-attention do not introduce any parameters and more importantly, can be applied to other document-level seq2seq tasks. 

\subsection{Position Encoding for Transformer} Position information plays a crucial role in Transformer to model the word order of the input sequence. The absolute position information~\cite{vaswani_etal:17,ke2021rethinking} is properly propagated to higher layers through residual connections. The relative positional encoding~\cite{shaw-etal-2018-self,qu-etal-2021-explore,ainslie-etal-2020-etc,huang-etal-2020-improve} extends the self-attention that can be used to incorporate relative position information among tokens within a sequence. Later, Transformer-based pre-trained models, including BERT~\cite{devlin:2019-bert} and BART~\cite{lewis-2020-bart} use fully learnable PEs. 
\citet{chen-etal-2021-simple} show the gains of the relative positional encoding coming from moving positional information from the input to attention. So \citet{chen-etal-2021-simple} propose the decoupled positional attention to encode position and segment information into the Transformer models. Different from the above studies, in this paper we find that position information has almost vanished in the output of the encoder when the input becomes long on small datasets. And this could be easily resolved by letting the attention modules be explicitly aware of the position information for query-key pairs. 

\section {Conclusion}

In this paper, we have investigated the main reasons for training failure on the document-to-document Transformer, especially on small datasets. Through the position information probing tasks, we find that position information in the source representation has almost vanished at the top layer.
Therefore, we propose a simple position-aware self-attention and cross-attention by explicitly adding the position embeddings to the query-key pairs in the attention function.
The proposed P-Transformer is a general, truly Doc2Doc model that directly translates a source document to its corresponding target document.
Experimental results on several datasets show that P-Transformer significantly outperforms the strong baseline and achieves state-of-the-art performance.

Though the P-Transformer achieves good performance on Doc2Doc translation. 
However, memory usage will be a bottleneck for documents with thousands of tokens. So the computation efficiency on long-range sequence processing in Doc2Doc translation is the limitation that remains to be further explored. Moreover, in the future, we will evaluate the proposed P-Transformer against other document-level seq2seq tasks, e.g., text summarization~\cite{bajaj-etal-2021-long}, text simplification~\cite{sun-etal-2021-document} or chat translation~\cite{liang-etal-2021-modeling}.

\footnotesize
\bibliographystyle{IEEEtranN}

\end{document}